\documentclass{article}

\usepackage[nonatbib, final]{nips_2018}

\usepackage[utf8]{inputenc} 
\usepackage[T1]{fontenc}    
\usepackage{hyperref}       
\usepackage{url}            
\usepackage{booktabs}       
\usepackage{amsfonts}       
\usepackage{nicefrac}       
\usepackage{microtype}      

\usepackage{subfiles}
\usepackage{graphicx, subfigure}
\usepackage{multirow}
\usepackage{array} 
\usepackage{color} 
\usepackage{nicefrac}
\usepackage{authblk}

\title{Adversarial Examples as \\ an Input-Fault Tolerance Problem}

\author[1,2]{Angus Galloway}
\author[3,4]{Anna Golubeva}
\author[1,2]{Graham W.~Taylor}
\affil[1]{School of Engineering, University of Guelph}
\affil[2]{Vector Institute for Artificial Intelligence}
\affil[3]{Department of Physics and Astronomy, University of Waterloo}
\affil[4]{Perimeter Institute for Theoretical Physics}

\def\biblio{\bibliographystyle{abbrv}\bibliography{../nips2018_zotero}}

\begin{document}
\def\biblio{}
\maketitle

\begin{abstract}


We analyze the adversarial examples problem in terms of a model's fault tolerance
with respect to its input.
Whereas previous work focuses on arbitrarily strict threat models, i.e.,
$\epsilon$-perturbations, we consider arbitrary valid inputs and
propose an information-based characteristic for evaluating tolerance
to diverse input faults.\footnote{Source available
at~\url{https://github.com/uoguelph-mlrg/nips18-secml-advex-input-fault}}

\end{abstract}

\section{Introduction}
\label{sec:intro}

Fault tolerance is a qualitative term that refers to the ability of a system to
perform within specification despite faults in its subsystems.
A way of characterizing a complex system's fault tolerance is
to measure its performance under induced faults of varying strength.
In particular for systems operating in safety-critical
settings~\cite{protzel1993performance},
it is desirable that the performance degrades gradually as a function of fault
severity and consistently so across a variety of fault types.

Most of the literature on the fault tolerance of artificial
neural networks considers~\emph{internal} faults,
such as deliberate~\cite{hinton1991lesioning, srivastava2014dropout} or
accidental~\cite{sequin1990fault, piuri2001analysis, tchernev2005investigating}
neuron outage.
Modern deep networks, however, are presented with increasingly complex
data, and many applications demand predictable performance~\emph{for all} inputs,
e.g., low-confidence outputs for out-of-distribution inputs.
Therefore, characterizing the fault tolerance of the overall system requires
considering the input itself as a source of~\emph{external}
faults~\cite{chandra2003fault, torres-huitzil2017fault}.
We suggest that the~\emph{adversarial examples} phenomenon, which
exposes unstable model behaviour for valid bounded inputs~\cite{szegedy2014intriguing,
nguyen2015deep, papernot2016limitations, carlini2017evaluating,
athalye2018obfuscated, anonymous2019adef}, be interpreted as a type of external
fault.

As a measure of a model's tolerance to adversarial attacks of increasing strength,
we propose the information conveyed about the target variable, i.e.,
the label in a classification task~\cite{shannon1948mathematical, tishby1999information}.
We find this measure to be more representative of a model's expected
robustness than a previous convention of reporting
the test error rate for fixed $\epsilon$-perturbations.\footnote{For demonstrations
as to how the latter approach can be incomplete c.f.~\cite{sharma2017attacking,
galloway2018adversarial, schott2018first, gilmer2018motivating, xiao2018spatially}.}
The proposed characteristic curves reduce the need for human supervision to
distinguish mistakes from ambiguous instances~\cite{song2018constructing,
brown2018unrestricted}, which is subjective and time consuming.

We expose a convolutional network to a range of different attacks
and observe that: i) the proposed robustness measure is sensitive to 
hyper-parameters such as weight decay, ii) prediction accuracies may be identical, 
while the information curves differ, and iii) more gradual changes in the 
information conveyed by the model prediction corresponds to improved adversarial
robustness.


\biblio 

\section{Methodology}
\label{sec:method}

We introduce a way of quantifying input-fault tolerance for arbitrary inputs
in terms of the mutual information (MI), $I(T; Y)$, between a model's categorical
prediction and the true label, represented by the random variables that $T$ and
$Y$, respectively. The MI can be written as
$I(T; Y) = H(Y) - H(Y | T)$, where $H$ denotes the Shannon entropy in bits,
which is a measure of uncertainty.
Perfect predictive performance is achieved when $H(Y | T)=0$, i.e.,
when there is no uncertainty about $Y$ given the
model's prediction $T$. The upper bound on $I(T; Y)$ is given by $H(Y)$,
which is 3.2 bits for the full (unbalanced) street-view house numbers
(SVHN)~\cite{netzer2011reading} test set. We use a random
sample of 1000 images from this set for our analysis.

For perturbation-based attacks, we plot MI versus the input
signal-to-noise ratio (SNR), defined as
$20\log_{10}\big(1 + \nicefrac{\|x\|_2}{\|\delta_x\|_2}\big)$ in dB,
for test inputs $x$, and noise $\delta_x$. The noise may be correlated with
$Y$ (adversarial perturbations) or uncorrelated (AWGN).
For vector field-based deformations~\cite{anonymous2019adef}, the maximum norm
of the vector field is used as a measure of perturbation strength instead, as it is
less clear how to standardize to SNR in this case. However, the choice of units
on the x-axis is not critical for the current analysis.

Datasets must be prepared identically for model comparison.
We suggest using the zero-mean and unit-variance standard,
which we implement with per-image mean subtraction, after
converting SVHN images from RGB to greyscale via the National Television System
Committee (NTSC) conversion.
Note that input signals with a non-zero mean, or DC bias, translate the curves along
the SNR axis if not removed; we provide additional reasons for preprocessing in
Appendix~\ref{sec:method-details}.

We subject the model to a broad array of faults: AWGN~\cite{cover1991elements,
chandra2003fault}, rotations and translations~\cite{engstrom2017rotation}, a
basic iterative method ``BIM''
($L_2$ and $L_\infty$ variants)~\cite{kurakin2017adversarial}, ``rubbish''
or ``fooling'' images~\cite{goodfellow2015explaining, nguyen2015deep}, and
deformations ``ADef''~\cite{anonymous2019adef}. 
This variety reflects real inputs that span a continuous range of signal
quality, and exposes defenses that mask
gradients~\cite{papernot2017practical, athalye2018obfuscated}, or fit a specific
set of inputs, e.g., a fixed $\epsilon-L_\infty$ box.


\biblio 

\section{Evaluation}
\label{sec:evaluation}

\newcommand{\Perp}{\perp \! \! \! \perp}

\begin{figure}[]
\centering
\subfigure[]{
\label{fig:ft-l2-wd}
\includegraphics[width=0.49\textwidth]{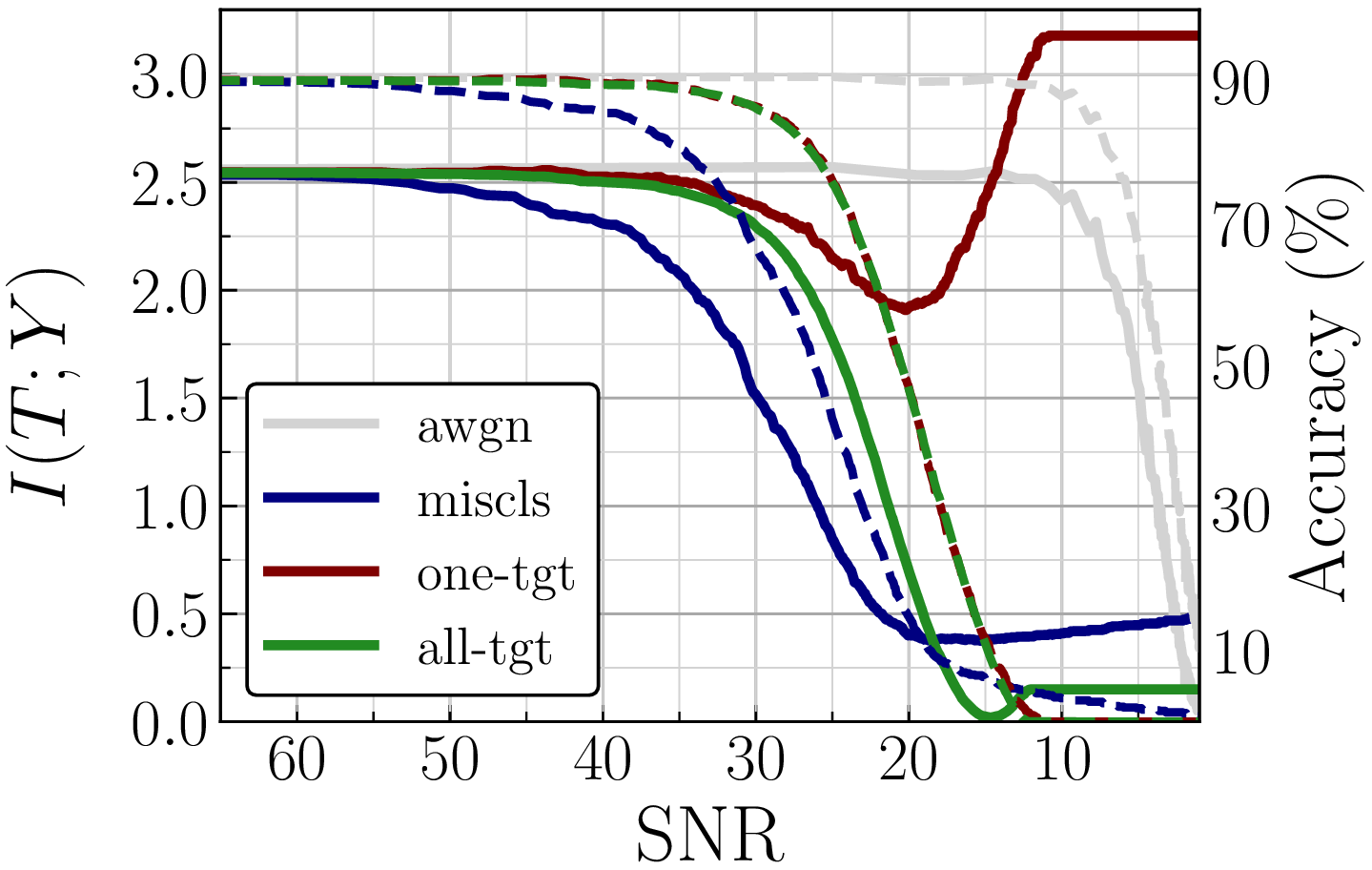}}
\subfigure[]{
\label{fig:ft-l2-no-wd}
\includegraphics[width=0.49\textwidth]{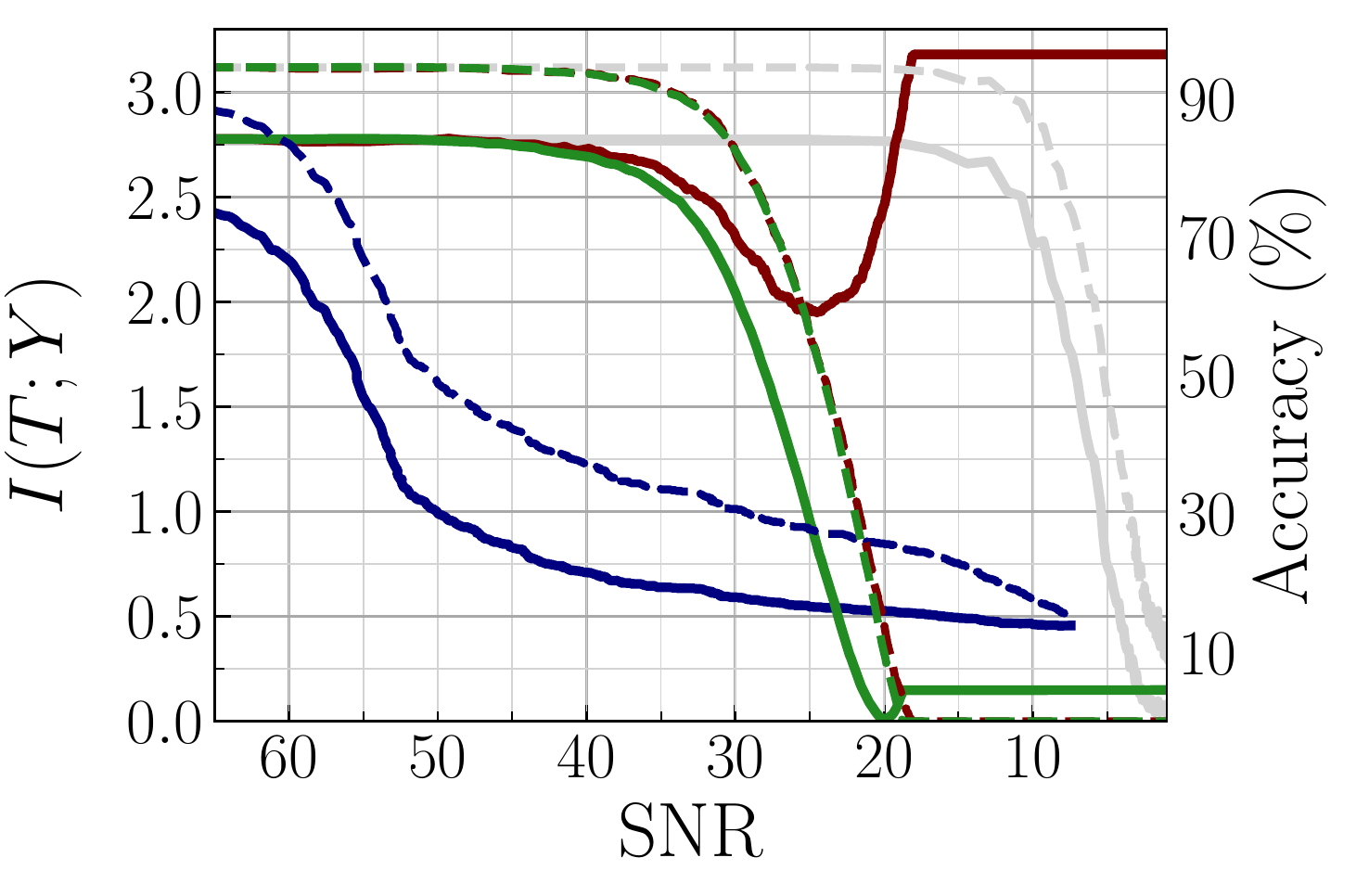}}
\caption{Motivating input-fault tolerance in terms of information conveyed about
the original label when models are subject to attacks with different
objectives. Model~\subref{fig:ft-l2-wd} was trained with weight decay,
and~\subref{fig:ft-l2-no-wd} without.
We report $I(T; Y)$ (solid line) and prediction accuracy (dashed line)
for a randomly selected subset of 1000 images from the
SVHN test set as a function of the SNR in dB for a BIM-$L_2$ attack with
three different misclassification objectives (details provided in the main text),
and AWGN for comparison. Note that for model~\subref{fig:ft-l2-no-wd}
gradients vanish before zero SNR can be reached.
}
\label{fig:motivate-ft-information}
\end{figure}

We begin by demonstrating why it is desirable to examine adversarial attacks
from the perspective of the information in the predictions rather than solely
evaluating prediction accuracy. Figure~\ref{fig:motivate-ft-information} shows
characteristic curves for two pre-trained models subject to BIM-$L_2$ attacks
with different adversarial objectives: misclassification ``miscls'',
which aims to make the prediction~\texttt{not} $Y$, and two targeted attack variants,
``one-tgt'', which maps each class label to a particular target label (we use 
the shift $0\rightarrow 1$, $1 \rightarrow 2$,~\dots, $9 \rightarrow 0$),
and ``all-tgt'', mapping each original label to each possible incorrect label. 
For comparison, we also show an additive white Gaussian noise (AWGN) perturbation.

The initial $I(T; Y)$ values for the clean test set are approximately $2.50$ and
$2.75$ bits, corresponding to prediction accuracies of $90\%$ and $94\%$ for
the model trained with weight decay~\ref{fig:ft-l2-wd} and
without~\ref{fig:ft-l2-no-wd}, respectively.
AWGN is the best-tolerated perturbation, as can be seen from the extended plateau
until very low SNR\@. Furthermore, it is the only one for which $I(T; Y)$ reduces to zero
along with the test accuracy. In general, $I(T; Y)$ declines initially
as the introduced perturbations cause mistakes in the model's prediction,
but it remains non-zero and behaves distinctly for the three adversarial objectives.

For instance, consider the ``one-tgt'' case: With increasing perturbation strength
the model maps inputs to the target label more consistently. The minimum
of this curve marks the transition point at which the perturbed input resembles
the target class more closely than its original class. Additional perturbations further refine
the input, such that the MI keeps increasing and reaches the upper
bound $I(T; Y) = H(Y)$.
That is, we observe perfect information transmission despite zero predictive
accuracy, indicating that the model's predictions are in fact correct -- the
input has been changed to an extent that it is a legitimate member of the
target class.

A similar effect occurs for the simpler misclassification case ``miscls'',
where in Figure~\ref{fig:ft-l2-wd} we observe a slow increase in $I(T; Y)$
for SNR~$\leq 20$, indicating that from this point on additional perturbations
systematically add structure to the input.
For the case where each wrong label is targeted ``all-tgt'', the MI vanishes
at the point of complete confusion, i.e., when the inputs are perturbed
to the extent that $H(Y|T) = H(Y)$, implying that the probability distribution
$p(Y|T)$ is uniform. Additional perturbations beyond this point reduce the
probability of the original class to zero, thus causing an ``overshoot'' effect
where $I(T; Y)$ increases to approximately $0.2$ bits, a final saturation value
that is independent of the model.

In general, note that targeted attacks require more degradation of the input
than a misclassification attack to achieve a desired performance drop; this is
reflected in the relative positions of the corresponding curves.

Next, we compare the characteristic MI curves and relate them to model robustness
for the case with weight decay, and a baseline without.
For the model with weight decay, presented in Figure~\ref{fig:ft-l2-wd},
$I(T; Y)$ is initially slightly lower, but both
the~\emph{decrease} and~\emph{increase} in $I(T; Y)$ for the ``one-tgt'' attack
are more gradual than the baseline in Figure~\ref{fig:ft-l2-no-wd}.
Furthermore, the gap between the initial and minimum value is smaller
in Figure~\ref{fig:ft-l2-wd} (approximately $0.50$ vs.~$0.75$ bits).

\begin{figure}[]
\begin{tabular}{m{0mm}c}
& \multirow{3}{*}{
\subfigure{
\includegraphics[width=0.96\textwidth]{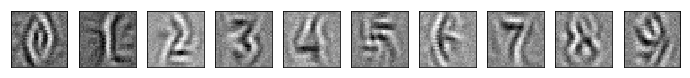} 
\label{fig:fool-wd}}} \\ \\
(a) & \\
& \\
& \multirow{3}{*}{
\subfigure{
\includegraphics[width=0.96\textwidth]{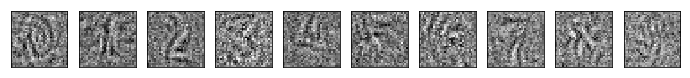} 
\label{fig:fool-no-wd}}} \\[3pt] \\
(b) & \\
&
\end{tabular}
{\caption{Adversarial examples classified with full confidence 100\% that were
initialized from white noise. Each of the respective digits are clearly
visible in~\subref{fig:fool-wd} for the more fault tolerant model (trained with
weight decay), while only some digits are faintly visible in~\subref{fig:fool-no-wd}.
Results for more SNR values are in Appendix~\ref{sec:advex}.}
\label{fig:svhn}}
\end{figure}

To further connect the gradual degradation property with qualitatively improved robustness
to adversarial examples, we use the BIM-$L_2$ method to craft
``fooling images''~\cite{nguyen2015deep} for each of the two models,
which are shown in Figure~\ref{fig:svhn}.
Starting from noise drawn from a Gaussian distribution with $\sigma=0.1$,
corresponding to an SNR of 20 dB w.r.t.~the original ($\sigma=1$) training data,
we applied BIM for each target label until full confidence was reached. 
The resulting images are very different:
The model with weight decay, which has a more gradual performance degradation,
yields images in Figure~\ref{fig:fool-wd} that emphasize the edge information
relevant to the digit recognition task defined by $Y$.
Conversely, the patterns for the baseline in Figure~\ref{fig:fool-no-wd} remain
contaminated
by noise, and do not reflect examples that would be identified by a human with
high confidence. Indeed, the model with weight decay classifies
the images in Figure~\ref{fig:fool-no-wd} with a mean margin of only 28\%,
while those of Figure~\ref{fig:fool-wd} are classified with full confidence by
the baseline model.

To summarize the analysis presented in Figure~\ref{fig:motivate-ft-information}: Test
accuracy consistently degrades to zero with
increasing attack strength for all adversarial objectives, while the MI does not.
This fact reflects the structure of the learnt clustering of the input space:
e.g., class 5 is transformed into an image resembling a ``2'' more often than it
becomes a ``1'' or a ``7'', indicating that the cluster corresponding to ``5'' is
closer to cluster ``2'' than to the others. The more predictable the alternative
incorrect predictions are, the more information $I(T; Y)$ is conveyed. Such
insights are lost by the accuracy.

\biblio 

\begin{figure}[]
\centering
\subfigure[]{
\label{fig:ft-adef}
\includegraphics[width=0.49\textwidth]{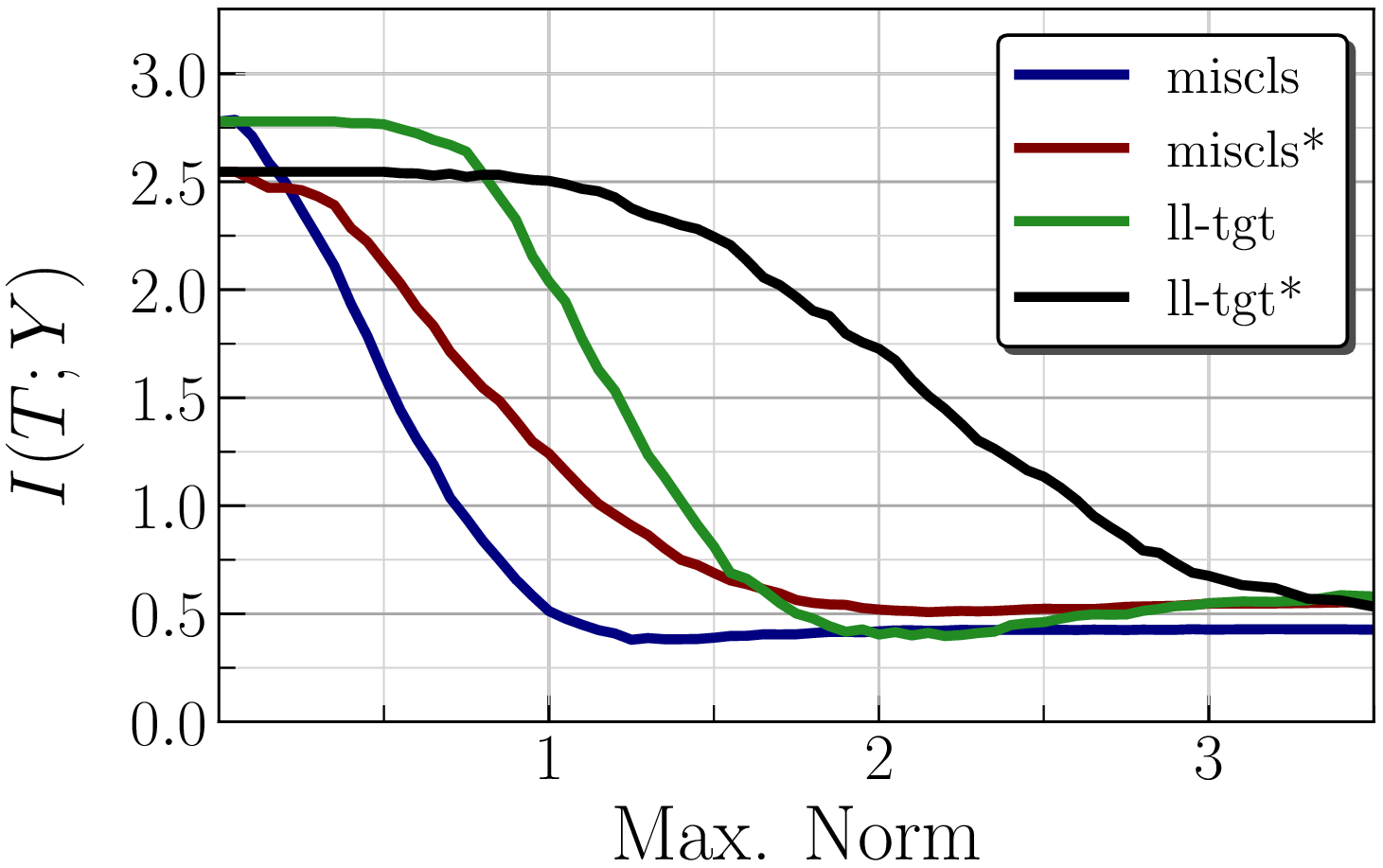}} 
\subfigure[]{
\label{fig:ft-rot-30}
\includegraphics[width=0.49\textwidth]{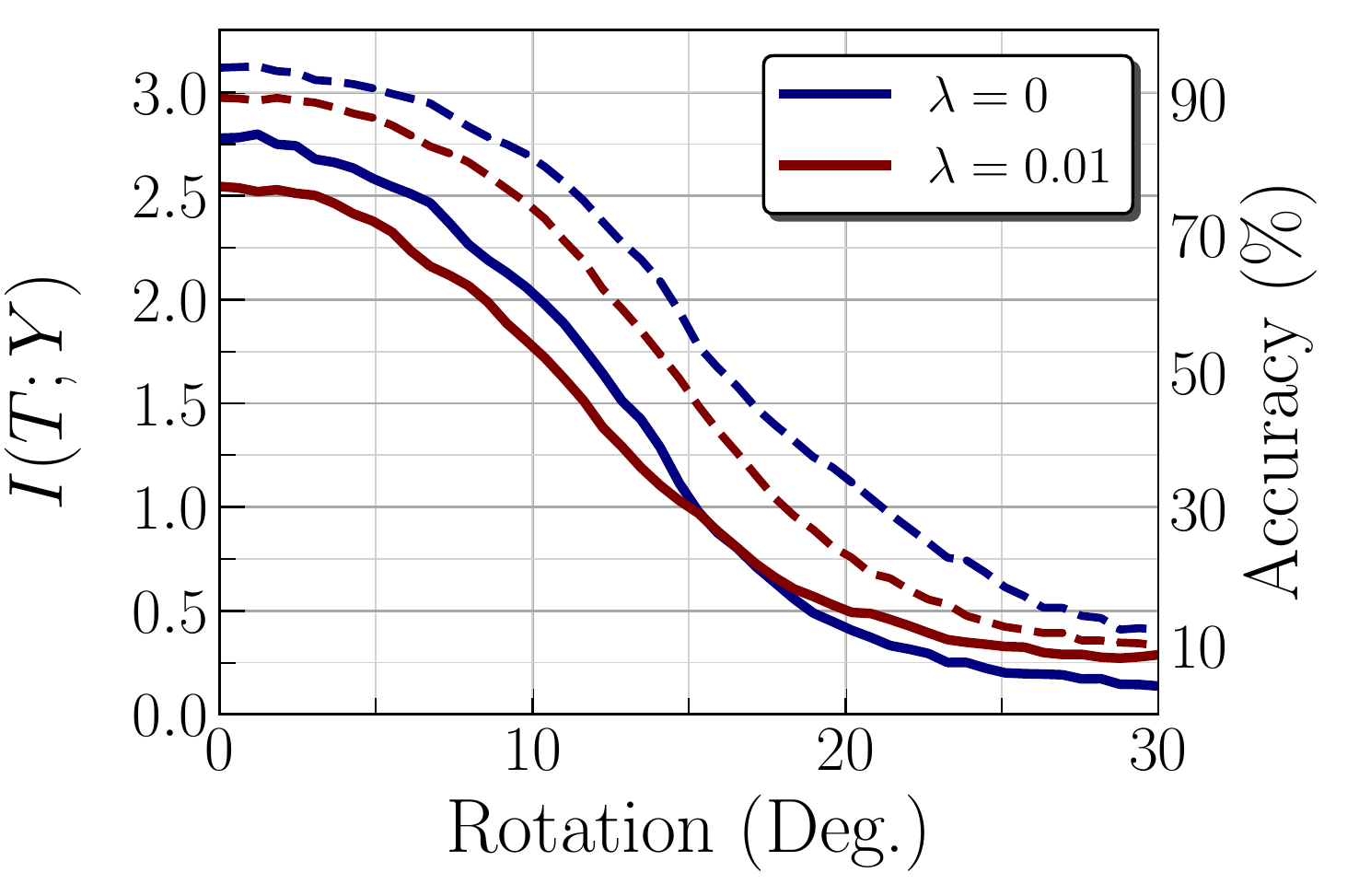}}
\caption{Tolerance to~\subref{fig:ft-adef} deformations produced by
``ADef'' for misclassification (``miscls'') and targeting of the
least likely label (``ll-tgt''), with accuracy omitted for
clarity, and~\subref{fig:ft-rot-30} rotations. Models
trained with weight decay denoted by ``*''.
Solid lines show the MI, dashed lines the prediction
accuracy. In~\subref{fig:ft-adef}, the maximum
permitted norm of the deforming vector field acts as a gauge for the attack
strength. In~\subref{fig:ft-rot-30}, we rotated images by $\leq 30^{\circ}$, as
per~\cite{engstrom2017rotation}; beyond this, a significant
fraction of the informative pixels were rotated out of the frame and digits
were far from their natural orientation.}
\label{fig:ft-spatial}
\end{figure}

Similar trends are observed under spatial attacks, including rotations
and deformations, as shown in Figure~\ref{fig:ft-spatial}.
Due to the additional digits on the canvas that distract from the center digit
for SVHN, we expect rotations and translations to be legitimately confusing for
models trained on this dataset.
We evaluate the model for $\pm 30^{\circ}$ rotations, as recommended
by~\cite{engstrom2017rotation} for $32\times32$ images, in Figure~\ref{fig:ft-rot-30}.
Translations must be handled with care again due to the peculiarities of the
SVHN dataset, where the only difference between otherwise identical images that
have a different label can be a translation of just a few pixels.

Finally, we consider attacks capable of producing arbitrary images via
deforming vector fields, rather than $L_p$ norm-based perturbations of the pixels.
The deformation algorithm, ``ADef''~\cite{anonymous2019adef}, builds
on the first-order DeepFool method~\cite{moosavi-dezfooli2016deepfool} to
construct smooth deformations
through vector fields that are penalized in terms of the supremum norm.
The behaviour observed in the tolerance curves of Figure~\ref{fig:ft-adef} aligns
well with results obtained for perturbation-based attacks, where again,
training with the weight decay constraint is most compelling, and targeted
attacks require greater changes to the input.
Several ADef examples and their predictions can be found in Appendix~\ref{sec:advex}.


\biblio 

\section{Conclusion}
\label{sec:conclusion}

We presented a new perspective connecting the adversarial examples problem
to fault tolerance --- the property that originally motivated the use
of neural networks in performance- and safety-critical settings.
We introduced a simple and intuitive measure for model tolerance:
information transmitted by the model under a given attack strength,
which is applicable across a diverse range of realistic fault models.
Adversarial examples show that although modern architectures may have
some inherent tolerance to internal faults, a combination of subtle design
principles and a thorough scope of the intended task are required
before they can demonstrate compelling tolerance to input faults.

\subsubsection*{Acknowledgments}

The authors wish to acknowledge the financial support of NSERC, CFI, CIFAR and
EPSRC\@. The authors also acknowledge hardware support from NVIDIA and Compute
Canada. Research at the Perimeter Institute is supported by the government of
Canada through Industry Canada and by the province of Ontario through the
Ministry of Research \& Innovation.


\bibliographystyle{abbrv}
\bibliography{nips2018_zotero}
\clearpage
\appendix

\section{Additional Detail Regarding the Methodology}
\label{sec:method-details}

It is essential that datasets be prepared identically for model comparison based
on the characteristic $I(T; Y)$ vs.~SNR curves.
As example, presence of a DC offset in the data, which commonly occurs in
natural images due to changes in brightness, will shift the SNR curves if not
corrected. We suggest adopting the zero-mean, unit-variance standard from image
processing, which we implement with per-image mean subtraction in this case for
SVHN, after first converting the RGB images to greyscale via NTSC conversion.

Generally speaking, preprocessing that helps with feature learning also
helps confer fault tolerance to adversarial attacks. Normally
one would also want to linearly decorrelate the pixels in the image, e.g., with
ZCA, but we found that the SNR was low enough in many of the SVHN images that
this eliminated low frequency gradient information essential for recognizing the
digit.

The adversarial examples literature generally leaves the
DC component in the dataset by simply normalizing inputs to [0, 1], which is
attractive from a simplicity perspective, and convenient for $\epsilon-L_\infty$
threat model comparisons, but we find that this practice itself contributes to
adverse model behaviour, such as excessively large prediction margins for purely
white noise patterns.

\biblio

\section{Additional Fault Tolerance Curves}
\label{sec:additional-fault}

\begin{figure}[] 
\centering
\subfigure[]{
\label{fig:ft-inf-wd}
\includegraphics[width=0.49\textwidth]{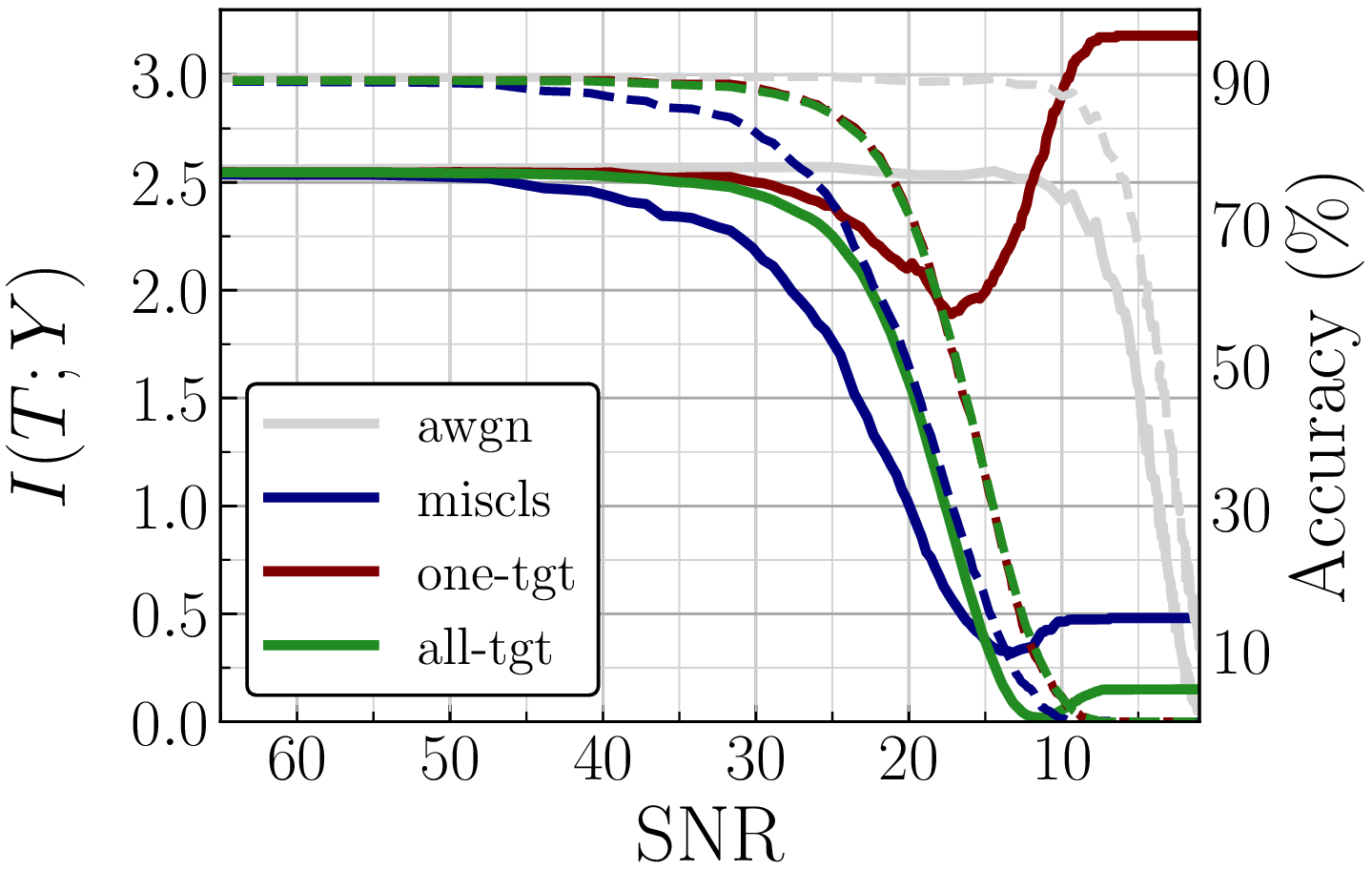}}
\subfigure[]{
\label{fig:ft-inf-no-wd}
\includegraphics[width=0.49\textwidth]{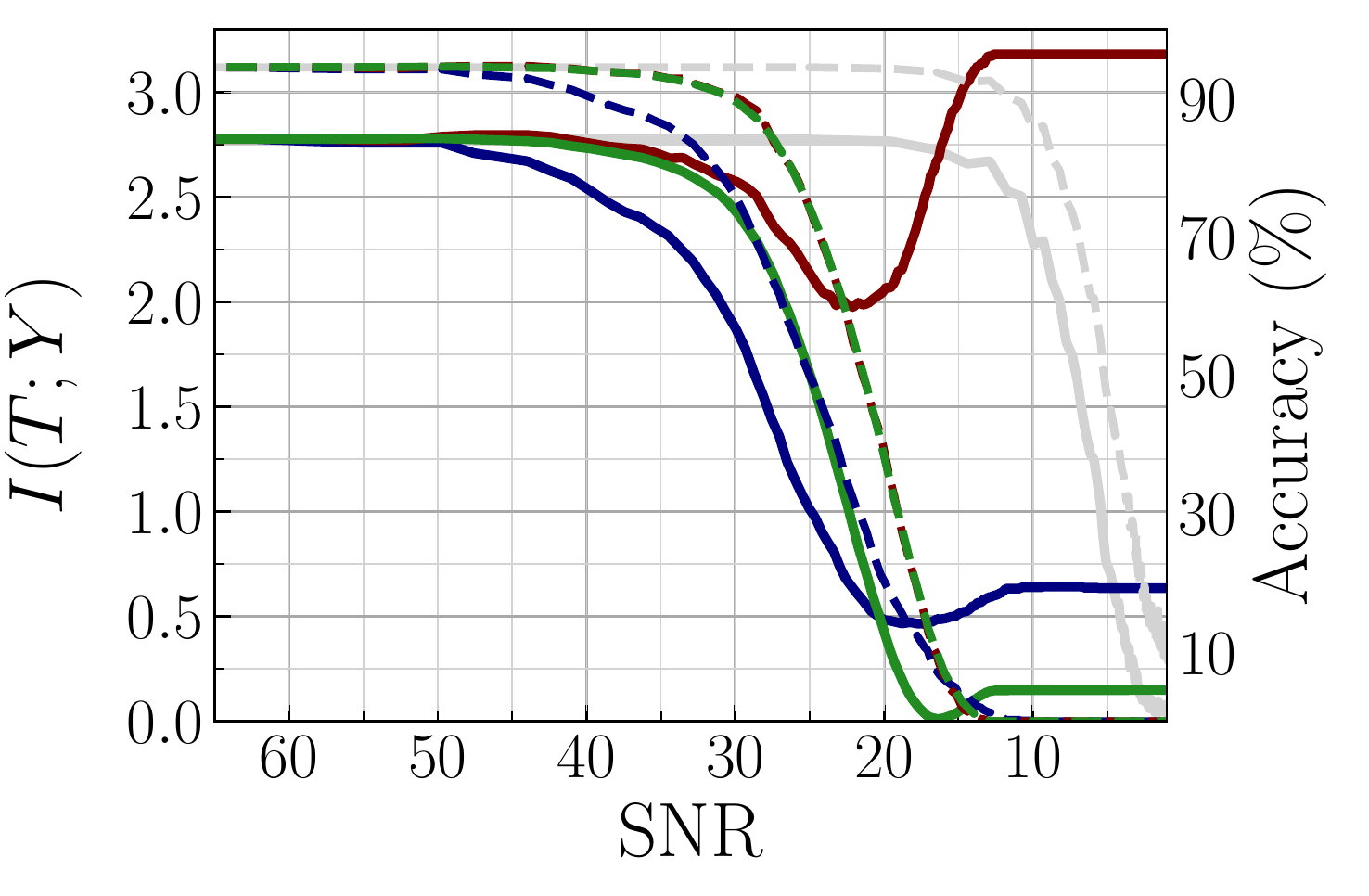}}
\caption{Input-fault tolerance in terms of information $I(T; Y)$ (solid line)
and accuracy (dashed line) for 1000 random samples from the SVHN test set, as a
function of the SNR for the
BIM-$L_\infty$ method with three different objectives (see main text for
details). As in Figure~\ref{fig:motivate-ft-information}, both models were
trained for 50 epochs with~\subref{fig:ft-inf-wd} weight decay,
and~\subref{fig:ft-inf-no-wd} without, with the former being more fault tolerant.}
\label{fig:ft-bim-inf}
\end{figure}

In Figure~\ref{fig:ft-bim-inf} we depict the same set of attacks as
in Section~\ref{sec:evaluation}, but for the $L_\infty$ variant of BIM\@.
Although Figures~\subref{fig:ft-inf-wd} and~\subref{fig:ft-inf-no-wd} appear
qualitatively similar, model~\subref{fig:ft-inf-wd}
trained with weight decay is shifted to the right.
By picking SNR values in the range 20--30 and moving upward until
intersection with the curves, we see that the degradation is more gradual
in Figure~\subref{fig:ft-inf-wd}.

\begin{figure}[] 
\centering
\subfigure[]{
\label{fig:ft-wd-l2-vs-linf}
\includegraphics[width=0.49\textwidth]{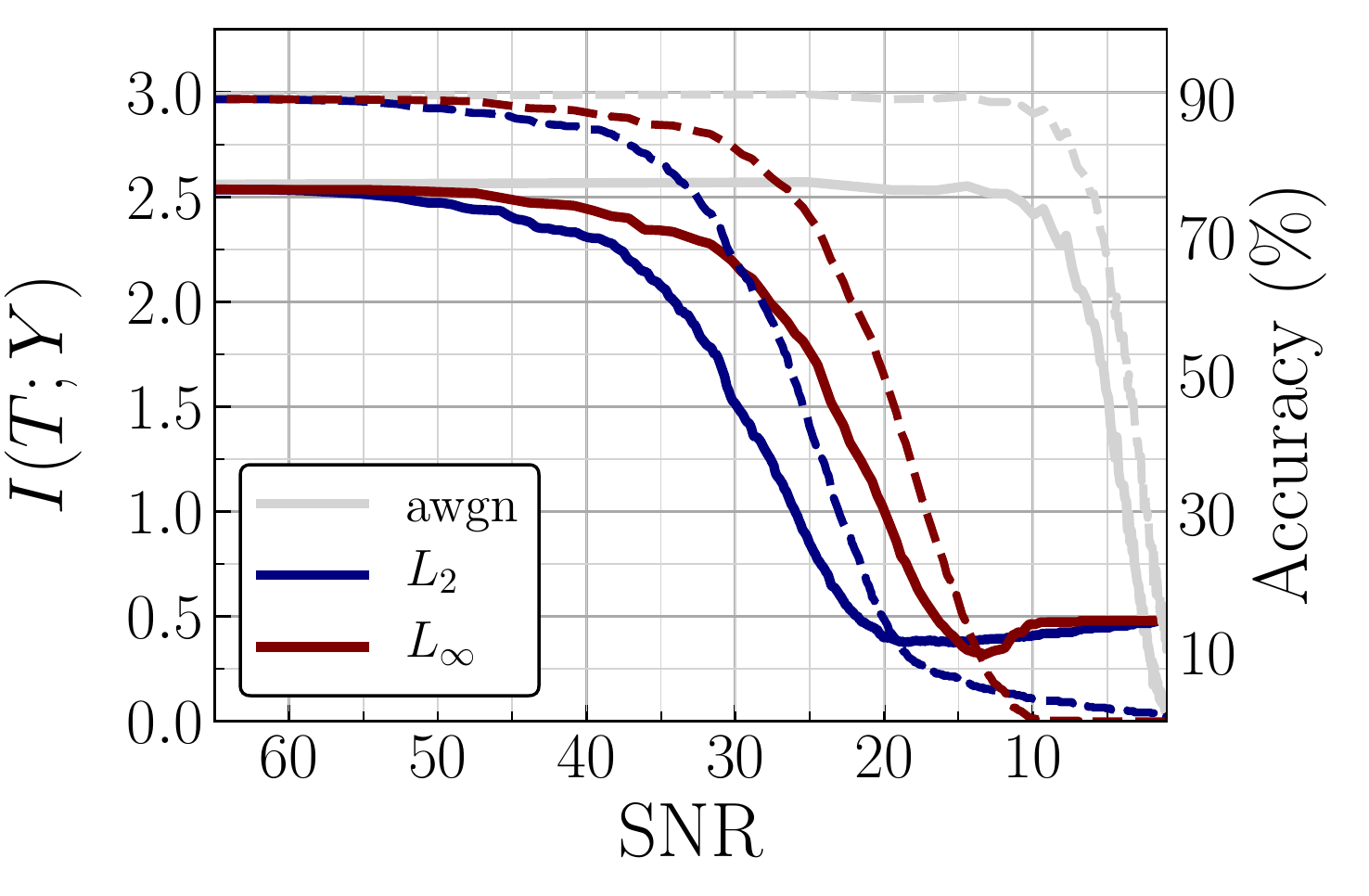}}
\subfigure[]{
\label{fig:ft-no-wd-l2-vs-linf}
\includegraphics[width=0.49\textwidth]{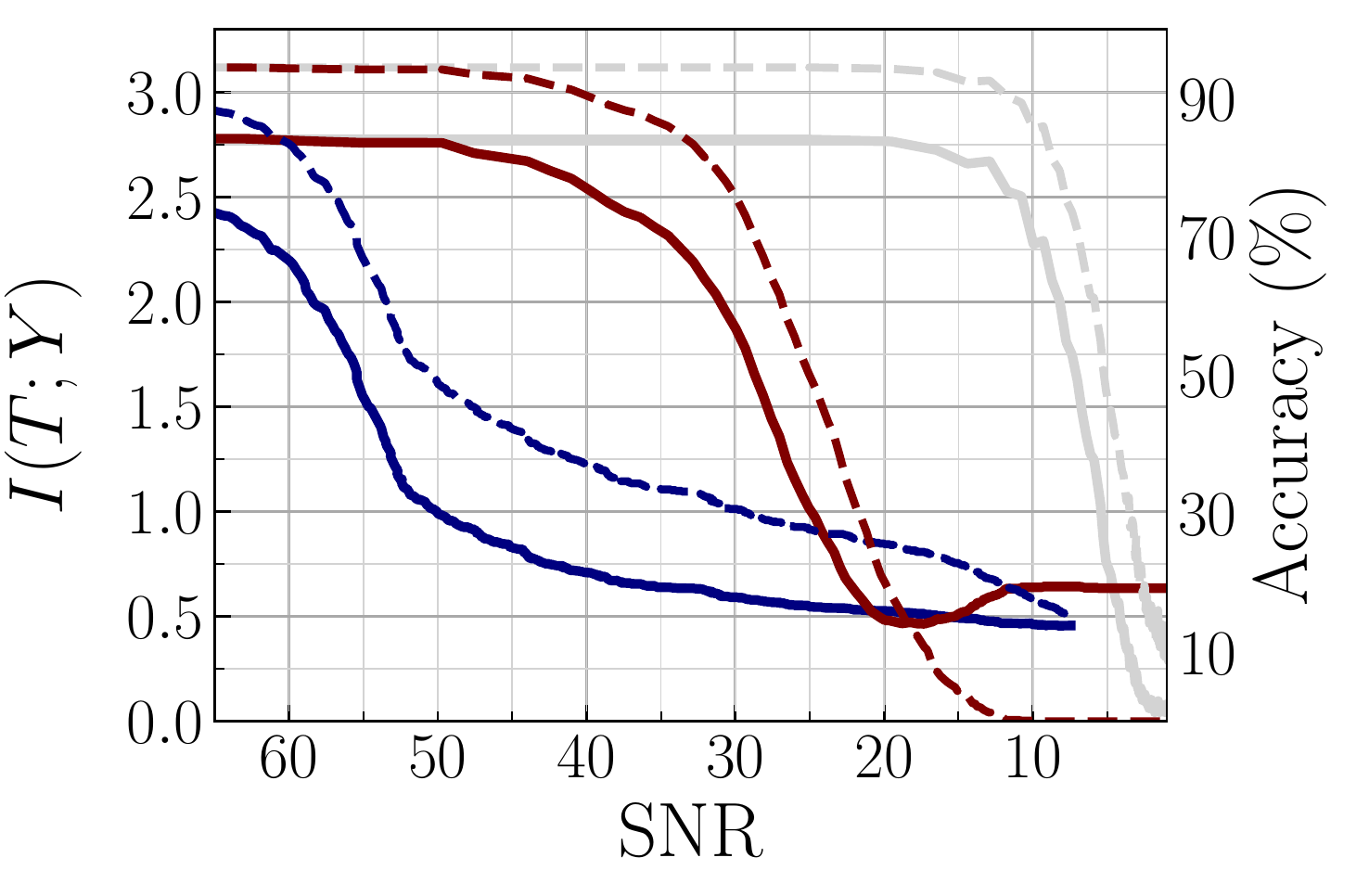}}
\caption{Comparison of BIM-$L_2$ and -$L_\infty$ misclassification attacks for
models trained with~\subref{fig:ft-wd-l2-vs-linf} weight decay
and~\subref{fig:ft-no-wd-l2-vs-linf} without. The $L_\infty$ variant is less
efficient for non-linear models, and thus requires more distortion for a given
performance level and is shifted to the right of the $L_2$ method.}
\label{fig:ft-bim-l2-vs-linf}
\end{figure}

In Figure~\ref{fig:ft-bim-l2-vs-linf} we show that curves for the $L_2$-BIM
adversary are generally to the left of those for the
$L_\infty$-BIM variant. This is expected since the $L_\infty$ constraint results
in a less efficient adversary for non-linear models.

\biblio

\section{Model Architecture}
\label{sec:arch}

We use a basic model with four layers, ReLU units, and a Gaussian
parameter initialization scheme. Unless specified otherwise, models were
trained for 50 epochs with inverse frequency class weighting, constant learning
rate (1e-2) SGD with the $L_2$ weight decay regularization constant $\lambda$
set to 1e-2 if weight decay is used, and a batch size of 128.
We summarize this model in Table~\ref{tab:cnn}, and respectively denote $h$, $w$, $c_{in}$,
$c_{out}$, $s$ as convolution kernel height, width, number of input and output
channels w.r.t.~each layer, and stride.



\begin{table}[h]
\centering
\caption{Fully-convolutional architecture adapted from the CleverHans library
tutorials~\cite{papernot2018cleverhans}, with 32 filters in first layer instead
of 64.}
\label{tab:cnn}
\begin{tabular}{ccccccr} 
Layer & $h$ & $w$ & $c_{in}$ & $c_{out}$ & $s$ & params \\
\toprule
\texttt{Conv1} & 8 & 8 & 1 & 32 & 2 & 2.0k\\ \hline 
\texttt{Conv2} & 6 & 6 & 32 & 64 & 2 & 73.8k\\ \hline 
\texttt{Conv3} & 5 & 5 & 64 & 64 & 1 & 102.4k\\ \hline 
\texttt{Fc} & 1 & 1 & 256 & 10 & 1 & 2.6k\\
\bottomrule
Total & -- & -- & -- & -- &  -- & \textbf{180.9k}\\
\end{tabular}
\end{table}

\biblio 

\section{Adversarial Examples}
\label{sec:advex}

\newcommand{\spatial}{0.95}

\begin{figure}[]
\begin{tabular}{m{0mm}c}
& \multirow{3}{*}{
\subfigure{
\includegraphics[width=0.96\textwidth]{img/fool/final_model_eps_50_seed_1_wd_0e+00_bim-ord2-100itr_gauss-u0e+00-std1e-01} 
\label{fig:fool-no-wd-snr-20}}} \\ [3pt] \\
(a) & \\
& \\
& \multirow{3}{*}{
\subfigure{
\includegraphics[width=0.96\textwidth]{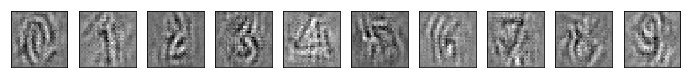} 
\label{fig:fool-no-wd-snr-40}}} \\ \\
(b) & \\
& \\
& \multirow{3}{*}{
\subfigure{
\includegraphics[width=0.96\textwidth]{img/fool/final_model_eps_50_seed_1_wd_1e-02_bim-ord2-100itr_gauss-u0e+00-std1e-01} 
\label{fig:fool-wd-snr-20}}} \\ \\
(c) & \\
& \\
& \multirow{3}{*}{
\subfigure{
\includegraphics[width=0.96\textwidth]{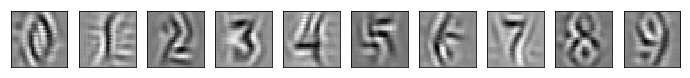} 
\label{fig:fool-wd-snr-40}}} \\ \\
(d) & \\
&
\end{tabular}
{\caption{Adversarial ``fooling images'' classified by a 100\% margin that were
initialized from white Gaussian noise with $\sigma$=1e-1, or an SNR of
20dB [\subref{fig:fool-no-wd-snr-20}--\subref{fig:fool-wd-snr-20}], and 1e-2,
or an SNR of 40dB [\subref{fig:fool-no-wd-snr-40}--\subref{fig:fool-wd-snr-40}]
w.r.t.~the original training data. Rows~\subref{fig:fool-wd-snr-20}
and~\subref{fig:fool-wd-snr-40} are for the model whose performance degrades
more gradually, trained with weight decay.}
\label{fig:fool-extra}}
\end{figure}

In Figure~\ref{fig:fool-extra} we show additional fooling images initialized
from noise of varying $\sigma$ or power. For a given value of $\sigma$, the
model trained with weight decay yields cleaner images with less task-irrelevant
noise.

\begin{figure}[]
\centering
\includegraphics[width=\spatial\textwidth]{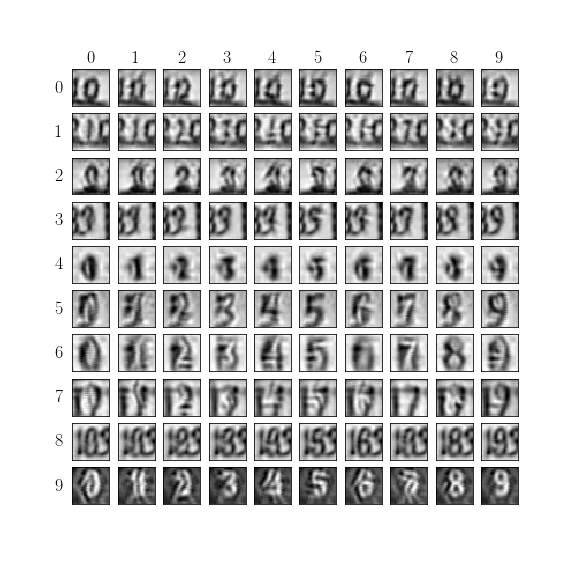}
\caption{Targeted adversarial examples obtained after running BIM-$L_2$ for the
least number of iterations sufficient to yield a 91\% mean margin.
The natural images are enumerated along the diagonal, with the source digit
class enumerated vertically, and the target class horizontally.}
\label{fig:bim-l2-advex-grid}
\end{figure}

In Figure~\ref{fig:bim-l2-advex-grid} we visualize targeted adversarial examples
constructed with the BIM-$L_2$ approach. In almost all cases, the features of
the source digit are manipulated to an extent such that a human observer would
likely agree with the target label.

\begin{figure}[]
\centering
\includegraphics[width=.9\textwidth]{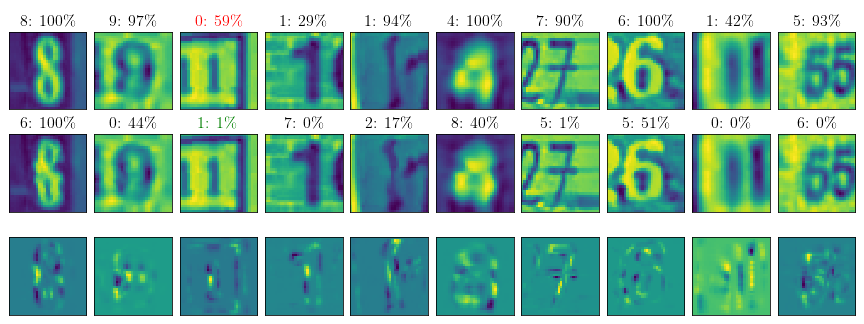}
\caption{Adversarial deformations constructed with the ADef algorithm,
with all incorrect labels as candidates, and max $T$-norm=4, for the model
with weight decay.
Above each image is the predicted label and prediction margin, defined as the
difference between the highest and second-highest softmax values of the prediction.
All deformations are concentrated
on the edges of the center digit, which is the most relevant information in
the image w.r.t.~the label. An ``8'' is deformed into a legitimate ``6'', and
other instances are in fact ambiguous -- e.g., a ``1'' on a high contrast background
could be a ``0'' with square edges, and ``5'' is relatively similar to ``6''  --
and are classified by a suitably small margin.}
\label{fig:adef-examples}
\end{figure}

In Figure~\ref{fig:adef-examples} we supplement the information curves for
adversarial deformations, ``ADef'', by showing qualitative examples,
complete with reasonably interpretable deformations. These examples were not
cherry-picked -- we arbitrarily sliced a set of ten examples from the test set,
and the misclassification confidence was either low, or in the case where
examples were misclassified with high confidence, they were usually changed into
the target class, e.g., an ``8'' deformed into a legitimate ``6''. It is possible
that a different attack, e.g.~``stAdv'', may find higher-confidence misclassifications,
but these results are nonetheless encouraging and show how attack success rates
(100\% in Figure~\ref{fig:adef-examples}) can lead to a false sense of
vulnerability.


\section{The Test Set ``Attack''}
\label{sec:test-time-attack}

Our solution achieving roughly 90.4$\pm$0.2\% clean test accuracy for the SVHN dataset
misses the mark in terms of state-of-the-art results, e.g.,
98.70$\pm$0.03\%~\cite{devries2017improved}.\footnote{Both
results averaged over five runs.}
It was recently suggested that methods which increase the error rate on the test
set are more vulnerable to a hypothetical ``test set attack'', in which an
attacker benefits simply by virtue of typical inputs being misclassified more
often~\cite{gilmer2018motivating}. Does such a reduction in test accuracy imply
the model is less secure?

Test accuracy is an application-specific constraint, or criteria,
which is~\emph{known} during the design phase. Not only can this be communicated
in advance to users of the model, it can also be controlled, e.g.,~by collecting
more data, as suggested by~\cite{schmidt2018adversarially}. Adversarial examples
characterize a situation in which the designer lacks such control for
the~\emph{overwhelming majority of valid inputs}, i.e., adversarial subspaces are not
usually rare. Such control can be reclaimed by demonstrating fault-tolerance for
attacks previously unseen to the model.

Although it is obvious that we should avoid unnecessarily limiting test accuracy,
in performance-critical settings we are primarily concerned with behaviour that
differs during deployment from that which was observed during the design phase.
A model that cannot achieve sufficiently high accuracy should not be
deployed in a security sensitive application in the first place, whereas high
accuracy on a subset of inputs could lead to a false sense of security, and
irrecoverable damages if a product is deployed prematurely.

It is crucial that we communicate precisely what our model does, i.e., is it
expected to recognize cars, trucks, and animals~\emph{in general}, or only those
appearing in a similar context, and at a given distance from the camera as in a
particular database, e.g., the ``Tiny Images''~\cite{torralba200880}.
Recent work found test performance degradations of 4--10\% absolute accuracy
when natural images were drawn from the same database~\cite{recht2018cifar10}, such
a large discrepancy in claimed versus obtained performance could be unacceptable
in many benign settings, and calls into question the significance
of solely numerical improvements on the state-of-the-art. The outlook is likely
less promising for the more general recognition case.

The old adage ``garbage-in, garbage-out'' suggests that we
should be at least as rigorous in ensuring models are consistently fed high
quality data capable of revealing the intended relationship, as we are rigorous
in our threat models. Predicting ``birds'' versus ``bicycles'' with no confident
mistakes~\cite{brown2018unrestricted} could be difficult to learn from finite
data without being more specific about the problem, e.g., is the bird's whole
body in the frame? What is the approximate distance from the camera? Is the bird
facing the camera? Our model is expected to recognize any
house number depicted with an Arabic numeral from a typical ``street-view''
distance for the given (unknown) lens, and otherwise yield a low-confidence
prediction.

\biblio 

\end{document}